\documentclass[12pt]{article}
\usepackage{spconf,amsmath,graphicx}
\usepackage{amssymb,amsfonts,float,subfig,stackengine}
\usepackage{algorithmic}
\usepackage{textcomp}
\usepackage{xcolor}
\usepackage{booktabs}
\usepackage{multirow}

\title{Attack on Scene Flow using Point Clouds}
\name{Haniyeh Ehsani Oskouie$^{\star}$ \qquad Mohammad-Shahram Moin$^{\dagger}$ \qquad Shohreh Kasaei$^{\star}$ \thanks{Preprint submitted to IEEE MLSP 2024.}}

\address{$^{\star}$ Sharif University of Technology\\
\{haniyeh.ehsani, kasaei\}@sharif.edu \\
$^{\dagger}$ ICT Research Institute\\
moin@itrc.ac.ir}

\onecolumn

\begin{document}
%
\maketitle
\begin{abstract}
Deep neural networks have made significant advancements in accurately estimating scene flow using point clouds, which is vital for many applications like video analysis, action recognition, and navigation. The robustness of these techniques, however, remains a concern, particularly in the face of adversarial attacks that have been proven to deceive state-of-the-art deep neural networks in many domains. Surprisingly, the robustness of scene flow networks against such attacks has not been thoroughly investigated. To address this problem, the proposed approach aims to bridge this gap by introducing adversarial white-box attacks specifically tailored for scene flow networks. Experimental results show that the generated adversarial examples obtain up to 33.7 relative degradation in average end-point error on the KITTI and FlyingThings3D datasets. The study also reveals the significant impact that attacks targeting point clouds in only one dimension or color channel have on average end-point error. Analyzing the success and failure of these attacks on the scene flow networks and their 2D optical flow network variants shows a higher vulnerability for the optical flow networks. Code is available at https://github.com/aheldis/Attack-on-Scene-Flow-using-Point-Clouds.git.
\end{abstract}
\begin{keywords}
Scene Flow Estimation, White-box Adversarial Attacks, Point Clouds
\end{keywords}
\section{Introduction}
\label{sec:intro}

Scene flow estimation is a crucial task that involves comprehending movements and interrelationships of objects within a three-dimensional scene. Unlike basic optical flow, which focuses solely on motion of individual pixels, the scene flow estimation considers a broader-scale motion and spatial connections between objects. By estimating the scene flow, valuable insights are gained that can be applied in diverse applications such as object tracking, video analysis, action recognition, navigation, liveness detection, and authentication. 
In this problem, the inputs are point clouds (PC) from two consecutive frames. Each point in the point cloud has multiple features including but not limited to color and point coordinates. The position of a point $p_i$ is represented by
$(x_i, y_i, z_i)$.
A set of vectors $(u_i, v_i, w_i)$ is viewed as the scene flow where each vector corresponds to the displacement of a point in the point cloud. 
The objective for this problem is to determine the scene flow $(U, V, W)$ 
between the point clouds 
$PC1$ at time $t$ and 
$PC2$ at time $t + 1$, 
such that it minimizes average end-point error (AEPE).

Conventional approaches to the scene flow estimation face challenges when dealing with fast-moving objects, occlusions, scenes with low texture, or blurry surfaces. To address these limitations and enhance both accuracy and speed, researchers have turned to deep neural networks, leveraging their ability to learn intricate patterns from extensive datasets. However, these networks are not impervious to adversarial attacks, which pose a significant threat to the accuracy and reliability of scene flow estimation. To date, despite the widespread adoption of the deep neural networks in this domain, there has been limited research investigating the impact of adversarial attacks on these networks. This knowledge gap underscores the need for a deeper understanding of the vulnerabilities and potential consequences associated with such attacks in scene flow estimation. In this paper, the focus is on extending the adversarial white-box attacks to the scene flow neural networks. To achieve this goal, FGSM-SF and PGD-SF are proposed which are specifically designed for the scene flow networks. As investigated in the paper, the scene flow networks are vulnerable to the designated adversarial attacks. The findings also reveal that attacks targeting point clouds in either a single dimension or color channel have a substantial impact on the average end-point error. Additionally, by comparing the robustness of the scene flow networks with their 2D counterparts, a higher level of robustness is observed for the scene flow networks. 

\section{Literature Review}
\label{sec:literature-review}
\subsection{Scene Flow}
State-of-the-art deep learning approaches have replaced the traditional optimization process by directly utilizing convolutional networks to predict scene flow \cite{b2, b3, b4, b5, b7, b9}. Liu et al. introduce the first deep neural network, named FlowNet3D, for the scene flow estimation directly from the point clouds, focusing on hierarchical features and flow embeddings \cite{b4}. Puy et al. introduce FLOT for the scene flow estimation on the point clouds using optimal transport principles \cite{b3}. Several studies in the field have focused on enhancing the generalization ability of the models to improve their performance across diverse scenarios \cite{b2, b7}. More recently, Wu et al. introduce PointPWCNet, a deep neural network designed for supervised and self-supervised scene flow estimation on the 3D point clouds \cite{b9}.
While these studies mainly focus on developing deep neural networks for the scene flow estimation from the point clouds, they do not address the evaluation of the networks' robustness against adversarial attacks. Therefore, in this paper, adversarial attacks are applied to the aforementioned network to compare the robustness of different networks designed for the scene flow estimation.

\subsection{Adversarial Attacks}
Adversarial attacks in the context of deep learning networks involve intentionally perturbing the input data to deceive the networks into making incorrect predictions. These attacks exploit vulnerabilities in the decision-making process of the networks, leading to erroneous outputs. In certain scenarios, the generated adversarial examples using these methods are intentionally fed into the network as part of the process to improve its resilience against potential attacks \cite{b11}. Recent research has revealed that neural network classifiers are further susceptible to universal adversarial attacks, which vividly expose their inherent weaknesses \cite{b12, b13}. 

A study by Szegedy et al. is a significant contribution to the robustness evaluation of the neural networks. The authors introduce Fast Gradient Sign Method (FGSM), known for generating adversarial examples for classification tasks. By leveraging the linear nature of neural networks, FGSM efficiently computes perturbations in the input space to cause misclassifications. The study highlights the vulnerabilities of the neural networks to the adversarial perturbations and emphasizes the regularization effects of adversarial training \cite{b8}. Similarly, Madry et al. explore another method for designing the adversarial attacks on the deep neural networks, called projected gradient descent (PGD) as a universal "first-order adversary" to enhance the robustness. Their findings underscore the importance of defending against the adversarial examples, suggesting the feasibility of securing the neural networks \cite{b10}. 

Lately, there has been a growing focus on attacks targeting point clouds \cite{b6, pgd3d, attack}. Liu et al. are the first to extend the adversarial attacks to deep 3D point cloud classifiers, proposing new defenses that leverage the unique structure of 3D point clouds. Their research suggests that while 3D point cloud classifiers are vulnerable to the attacks, they can be more easily defended compared to 2D image classifiers \cite{b6}. More recently, Ranjan et al. have explored adversarial patch attacks on optical flow networks, highlighting their disruptive impact on the performance of the networks. They found encoder-decoder networks to be highly vulnerable, while spatial pyramid architectures exhibited greater robustness \cite{b1}.  

Building upon the insights gained from previous studies, the scene flow networks, which are closely related to optical flow networks, are potentially susceptible to similar adversarial attack approaches. Hence, investigating the effectiveness of the adversarial attacks on the scene flow networks would be a valuable extension of the existing research.

\begin{figure*}[h!]
    \centering
    \small
    \stackunder[15pt]{
    \stackunder[0pt]{\includegraphics[width=0.25\textwidth]{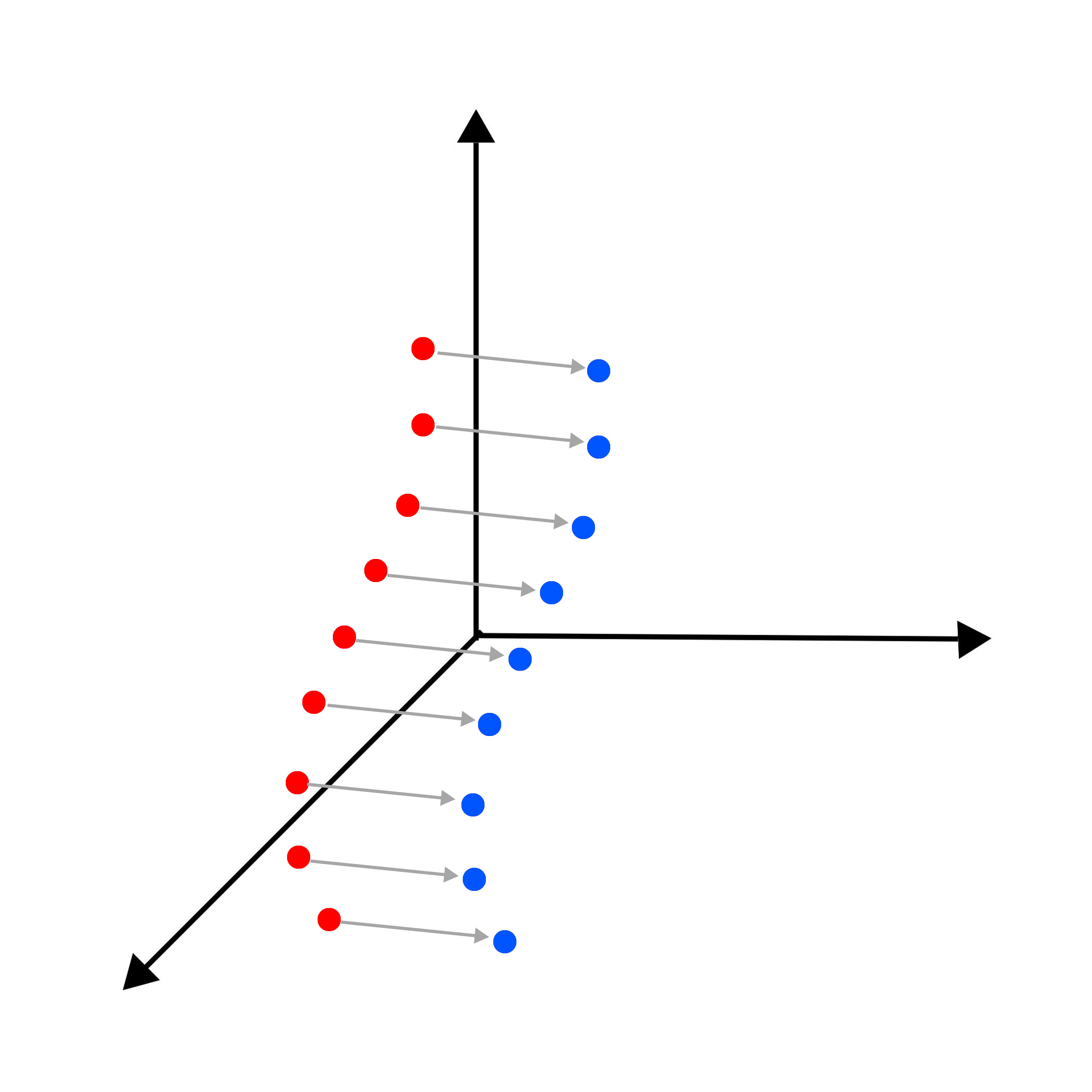}}{Before attack}
    \stackunder[0pt]{\includegraphics[width=0.25\textwidth]{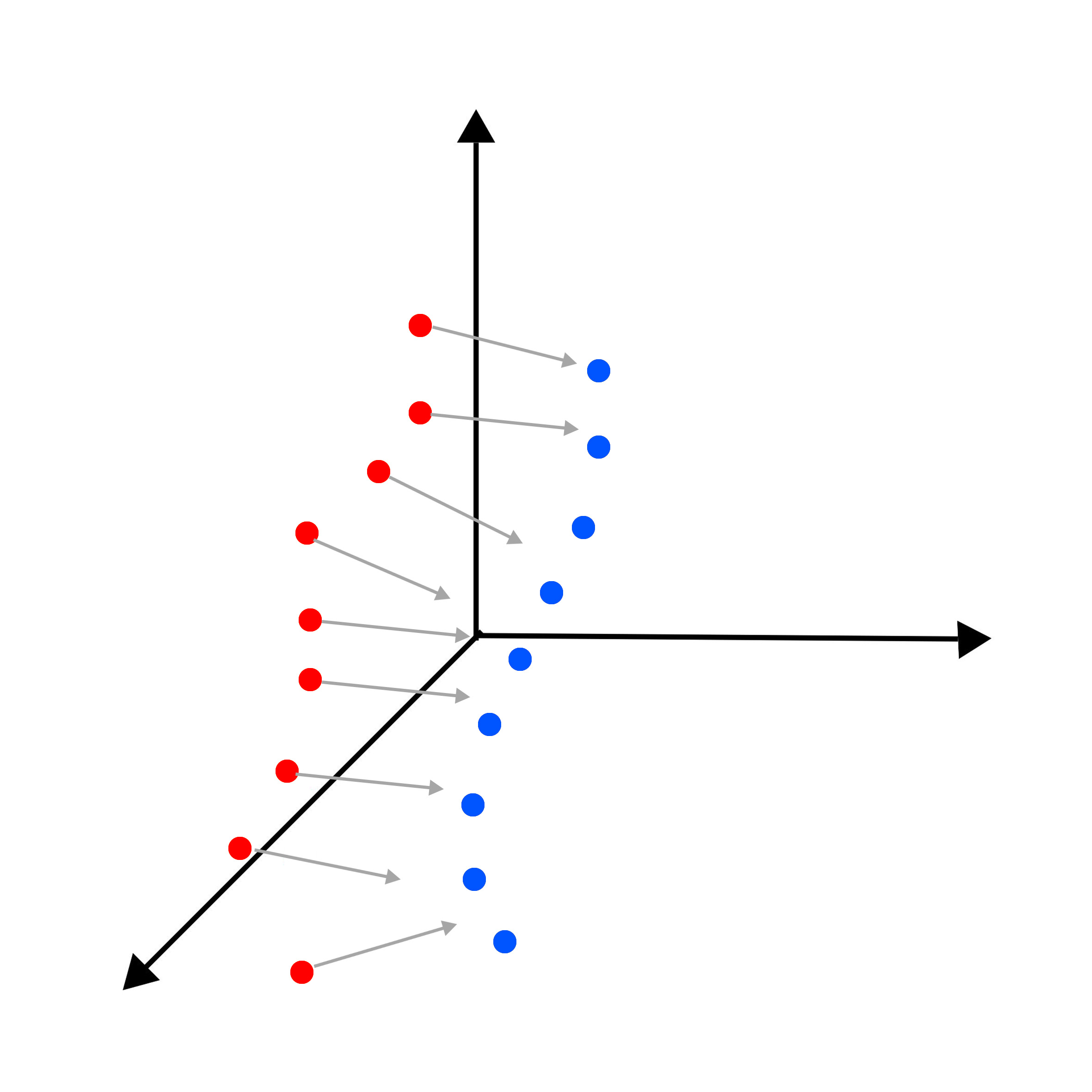}
    }{After attack}
    }{(a) Attack on dimensions}
    \hspace{5pt}
    \stackunder[15pt]{
    \stackunder[0pt]{\includegraphics[width=0.25\textwidth]{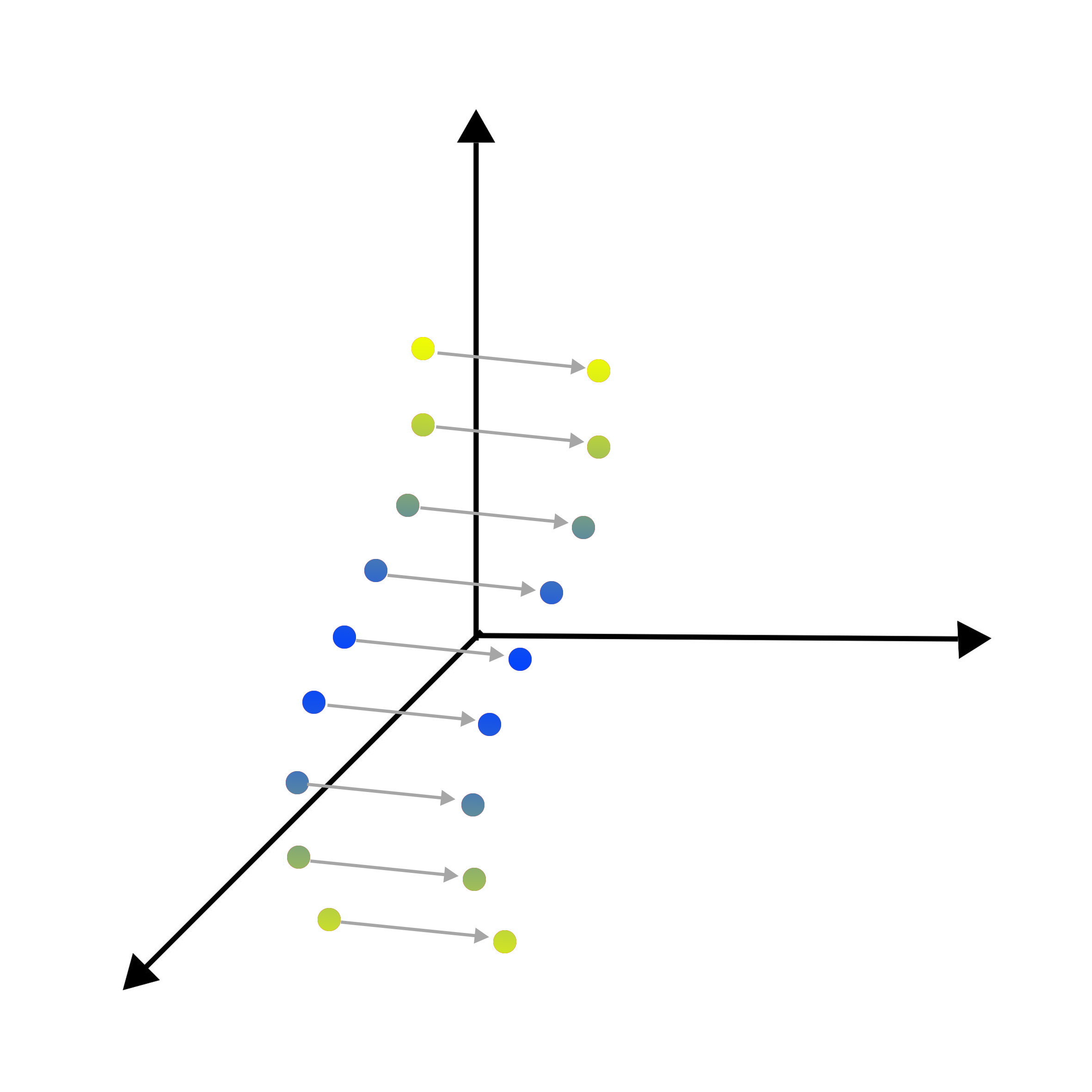}}{Before attack}
    \stackunder[0pt]{\includegraphics[width=0.25\textwidth]{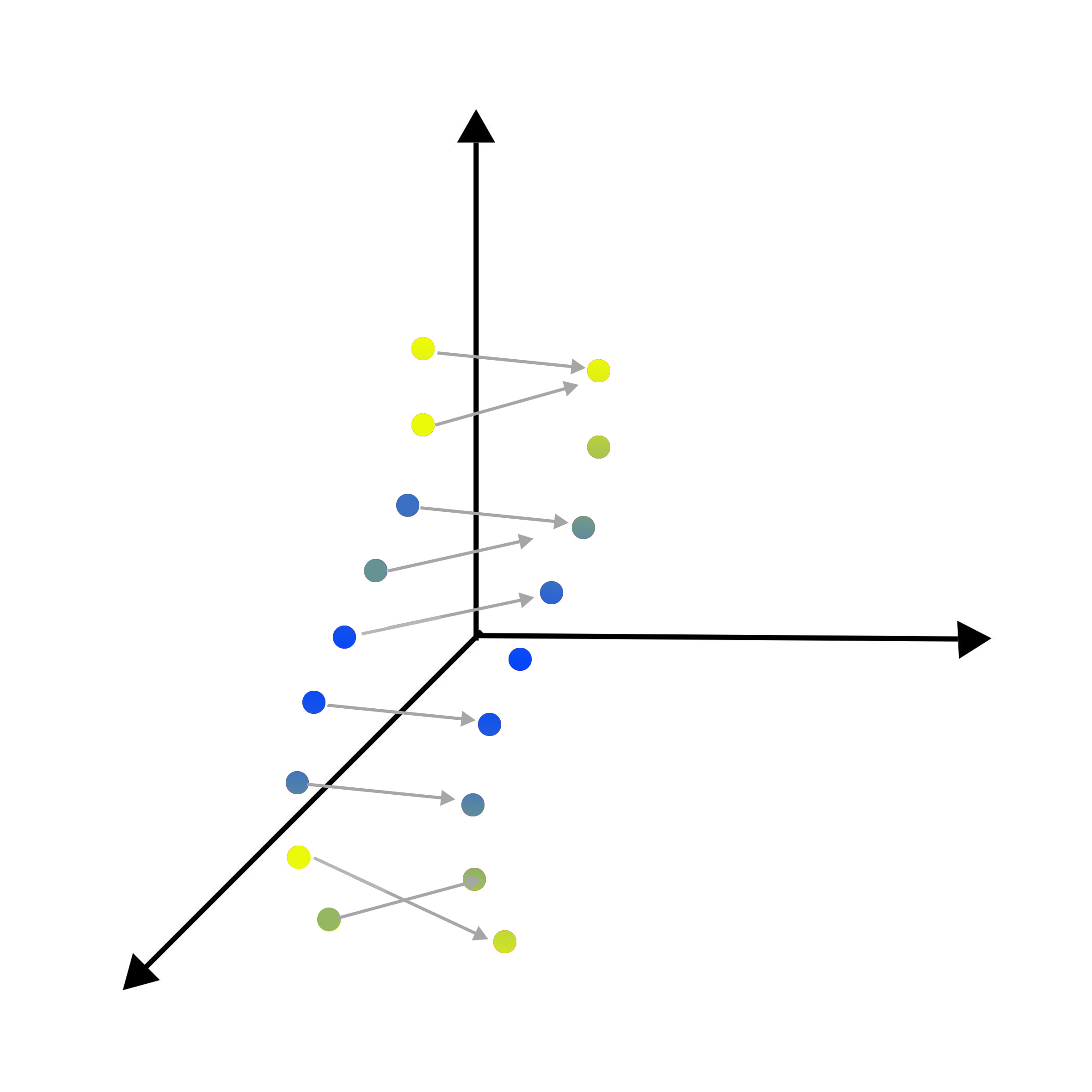}
    }{After attack}
    }{(b) Attack on color channels}
    \caption{
    Attacks on a sample of point cloud.
    The gray vectors illustrate scene flow. Perturbing the first point cloud results in significant alterations to the scene flow.
    }
    \label{Figure:Dimension_vs_color}
\end{figure*}

\section{Proposed Method}
\label{sec:method}

Adversarial attacks refer to deliberately modifying the input data in subtle ways to deceive neural networks, causing them to make predictions that deviate from true labels. The majority of scene flow datasets, which consist of the 3D motion of the objects in a scene, are either synthetically generated or limited, hindering the ability of the neural networks to generalize effectively to real-world scenarios. This challenge can be solved using FGSM \cite{b8} and PGD \cite{b10}, which allow adversarial examples to be generated during training phase, augmenting training data with perturbed samples, thus strengthening the models. Additionally, these examples can be considered as synthetic samples of distorted or occluded objects, effectively simulating the real-world scenarios where the objects may appear deformed. Utilizing these examples during training can aid the scene flow networks in improving their ability to handle such challenging cases, ultimately enhancing their overall performance.

Given a neural network $f$, $\epsilon$-norm bounded adversary $\delta$ for FGSM \cite{b6} and PGD \cite{pgd3d} is obtained by optimizing the following
\begin{equation}
\max_{\delta: \Vert\delta\Vert \le \epsilon} \ell \bigl(y,f(X+\delta) \bigr)
\end{equation}
where $\ell$ is the network's loss function, $X$ is the input, and $y$ is the ground-truth label of the input. 

For scene flow, the goal is to find a perturbation $\delta$ for the given point clouds $PC_1$ and $PC_2$ such that it maximizes the loss between the output and ground-truth scene flow $SF$. Here, end-point error (EPE) is considered as the main loss function. 
Therefore, the formula can be rewritten as
\begin{equation}
\max_{\delta: \Vert\delta\Vert \le \epsilon} \text{EPE} \bigl(SF, f(PC_1 + \delta, PC_2) \bigr).   
\end{equation}

Incorporating EPE as the primary loss function offers several advantages. First, it provides a reliable metric for comparing different models without the requirement of knowing their corresponding loss functions. Second, by disregarding terms related to regularization and occlusion, the vulnerabilities of the models can be specifically evaluated under normal circumstances. This approach aids in determining whether the models excel in handling typical scenarios.

For FGSM-SF, the $l_\infty$-bounded adversary $A$ is obtained by the following one-step scheme
\begin{align}
\text{L} = \text{EPE}(SF, f(PC_1, PC_2)) , \nonumber \\
A = PC_1 + \epsilon \cdot \text{sign}\bigl(\nabla_{PC_1} \text{L} \bigr).   
\end{align}

PGD-SF is a multi-step and stronger variant of FGSM-SF, and the adversarial example using this attack is calculated by
\begin{equation}
PC_1^{(t + 1)} = \Pi_{PC_1 + \mathcal{S}} \bigl( PC_1^{(t)} + \alpha \cdot \text{sign}(\nabla_{PC_1} \text{L} ) \bigr).   
\end{equation}
In the proposed method, it is important to highlight that the ground truth scene flow is not adjusted based on the perturbation applied to the input. The assumption is that the perturbation is minimal and does not alter the scene flow significantly.

Notably, the input data for scene flow contains either color or position information pertaining to the points. This distinction becomes significant when deciding which aspect to target for the attacks. If the input includes information about the color of the points, the choice is to target the color channels for the attacks; else, the attacks aim at the dimensions. Attacking the dimensions allows the generation of the point clouds that represent deformed objects. Conversely, attacking the color channels enables the simulation of local shadows and brightness variations, thereby enhancing the realism of the generated point clouds. An illustrative example showcasing these attacks can be observed in Fig. \ref{Figure:Dimension_vs_color}.

\begin{table*}
\begin{center}
\caption{Average end-point error of scene flow networks before and after FGSM-SF.}
\label{table:table1}
\resizebox{\textwidth}{!}{%
\begin{tabular}{l|l|c|cc|cc|cc|cc}
\toprule
\multirow{2}{*}{Dataset} & \multirow{2}{*}{Model} & Unattacked & \multicolumn{2}{|c|}{Attack on dimension 1} & \multicolumn{2}{|c|}{Attack on dimension 2} & \multicolumn{2}{|c|}{Attack on dimension 3} & \multicolumn{2}{|c}{Attack on all dimensions} \\
{} & {} & EPE & EPE & Rel & EPE & Rel & EPE & Rel & EPE & Rel \\
\midrule
\multirow{3}{*}{KITTI} & HPLFlowNet & 0.117 & 0.136 & +0.162 & 0.156 & +0.333 & 0.137 & +0.171 & 0.196 & +0.675 \\
{} & FLOT & 0.110 & 1.857 & +15.882 & 1.932 & +16.564 & 1.577 & +13.336 & 3.159 & +27.718 \\
{} & FlowNet3D & 0.299 & 1.426 & +3.769 & 1.603 & +4.361 & 1.588 & +4.311 & 2.932 & +8.806 \\
\midrule
\multirow{3}{*}{FlyingThings3D} & HPLFlowNet & 0.080 & 0.088 & +0.100 & 0.119 & +0.488 & 0.093 & +0.163 & 0.140 & +0.750 \\
{} & FLOT & 0.156 & 1.668 & +9.692 & 1.692 & +9.846 & 1.737 & +10.135 & 2.680 & +16.180 \\
{} & FlowNet3D & 0.155 & 1.532 & +8.884 & 1.576 & +9.168 & 1.631 & +9.523 & 2.555 & +15.484 \\
\bottomrule
\end{tabular}
}
\end{center}
\end{table*}

\begin{table*}
\begin{center}
\caption{Average end-point error of scene flow networks before and after PGD-SF.}
\label{table:table2}
\resizebox{\textwidth}{!}{%
\begin{tabular}{l|l|c|cc|cc|cc|cc}
\toprule
\multirow{2}{*}{Dataset} & \multirow{2}{*}{Model} & Unattacked & \multicolumn{2}{|c|}{Attack on dimension 1} & \multicolumn{2}{|c|}{Attack on dimension 2} & \multicolumn{2}{|c|}{Attack on dimension 3} & \multicolumn{2}{|c}{Attack on all dimensions} \\
{} & {} & AEPE & AEPE & Rel & AEPE & Rel & AEPE & Rel & AEPE & Rel \\
\midrule
\multirow{3}{*}{KITTI} & HPLFlowNet & 0.117 & 0.142 & +0.214 & 0.169 & +0.444 & 0.144 & 0.144 & 0.203 & +0.735 \\
{} & FLOT & 0.110 & 1.693 & +14.391 & 1.870 & +16.000 & 1.714 & +14.582 & 3.821 & +33.736 \\
{} & FlowNet3D & 0.299 & 1.690 & +4.652 & 1.746 & +4.839 & 2.020 & +5.756 & 3.141 & +9.505 \\
\midrule
\multirow{3}{*}{FlyingThings3D} & HPLFlowNet & 0.080 & 0.090 & +0.125	& 0.133 & +0.663 & 0.096 & +0.200 & 0.147 & +0.837 \\
{} & FLOT & 0.156 & 1.719 & +10.019 & 1.673 & +9.724 & 1.808 & +10.590 & 3.136 & +19.103 \\
{} & FlowNet3D & 0.155 & 1.544 & +8.961 & 1.479 & +8.542 & 1.683 & +9.858 & 2.341 & +14.103 \\
\bottomrule
\end{tabular}
}
\end{center}
\end{table*}

\section{Numerical Results}
\label{sec:results}
Numerical experiments were performed to evaluate the proposed FGSM-SF and PGD-SF
schemes on sampled point clouds from standard stereo image datasets including KITTI \cite{b14} and FlyingThings3D \cite{b15}. 
KITTI is a widely used benchmark dataset for computer vision tasks in autonomous driving. It comprises of real-world sensor data from LiDAR scans, along with annotations for tasks like object detection and scene understanding. FlyingThings3D is another dataset designed for scene flow estimation in aerial scenes. It contains around 32k stereo images with ground truth disparity and optical flow maps. 
Although KITTI is seemingly a better dataset to investigate for real-world scenarios, it provides only 200 training scenes which is insufficient for comprehensive training purposes. Hence, the models studied here are trained on FlyingThings3D and evaluated on both FlyingThings3D and KITTI.

\begin{figure*}[h!]
    \centering
    \small
    \stackunder[0pt]{\includegraphics[width=0.49\textwidth]{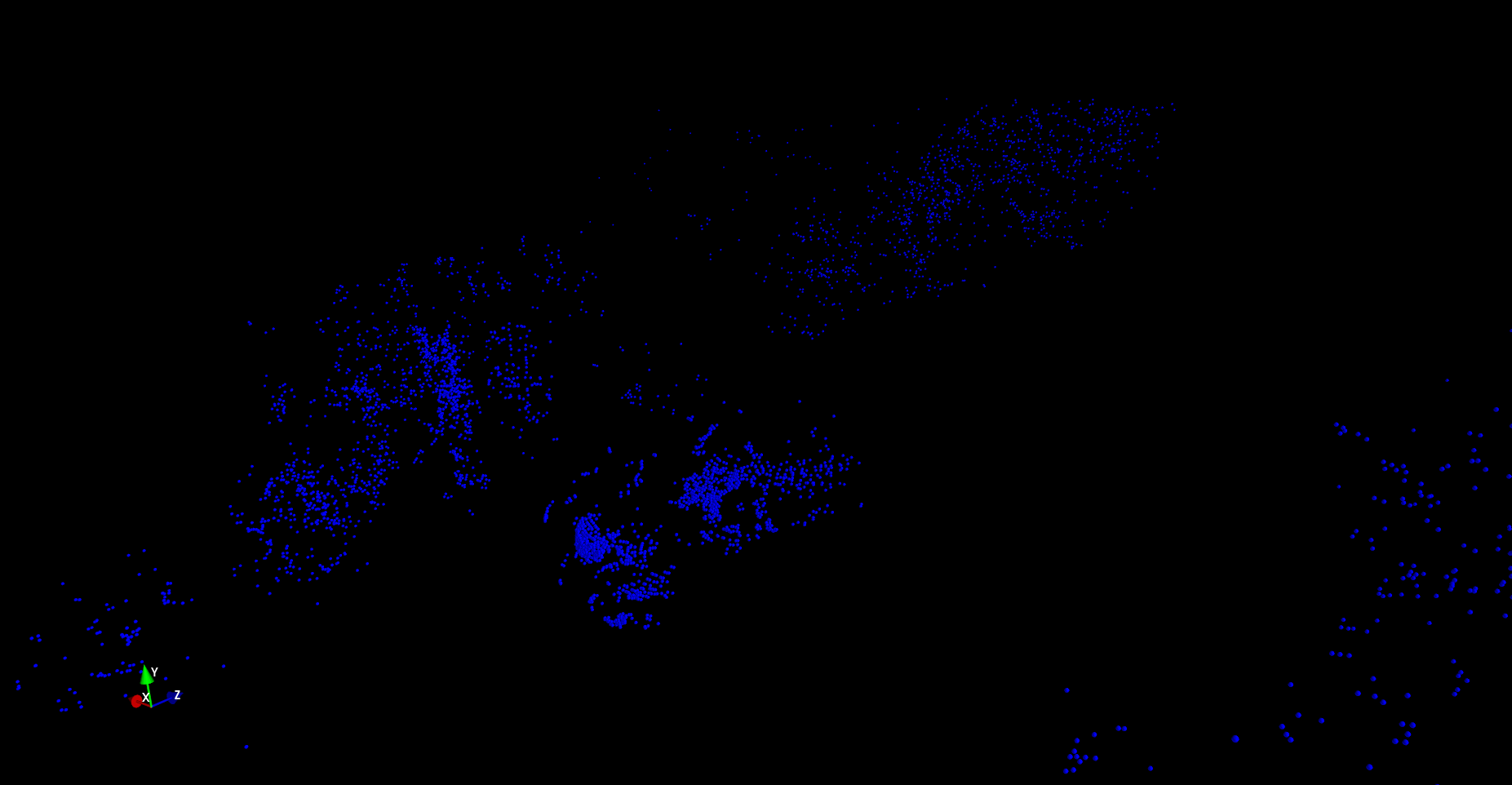}}{}
    \stackunder[0pt]{\includegraphics[width=0.49\textwidth]{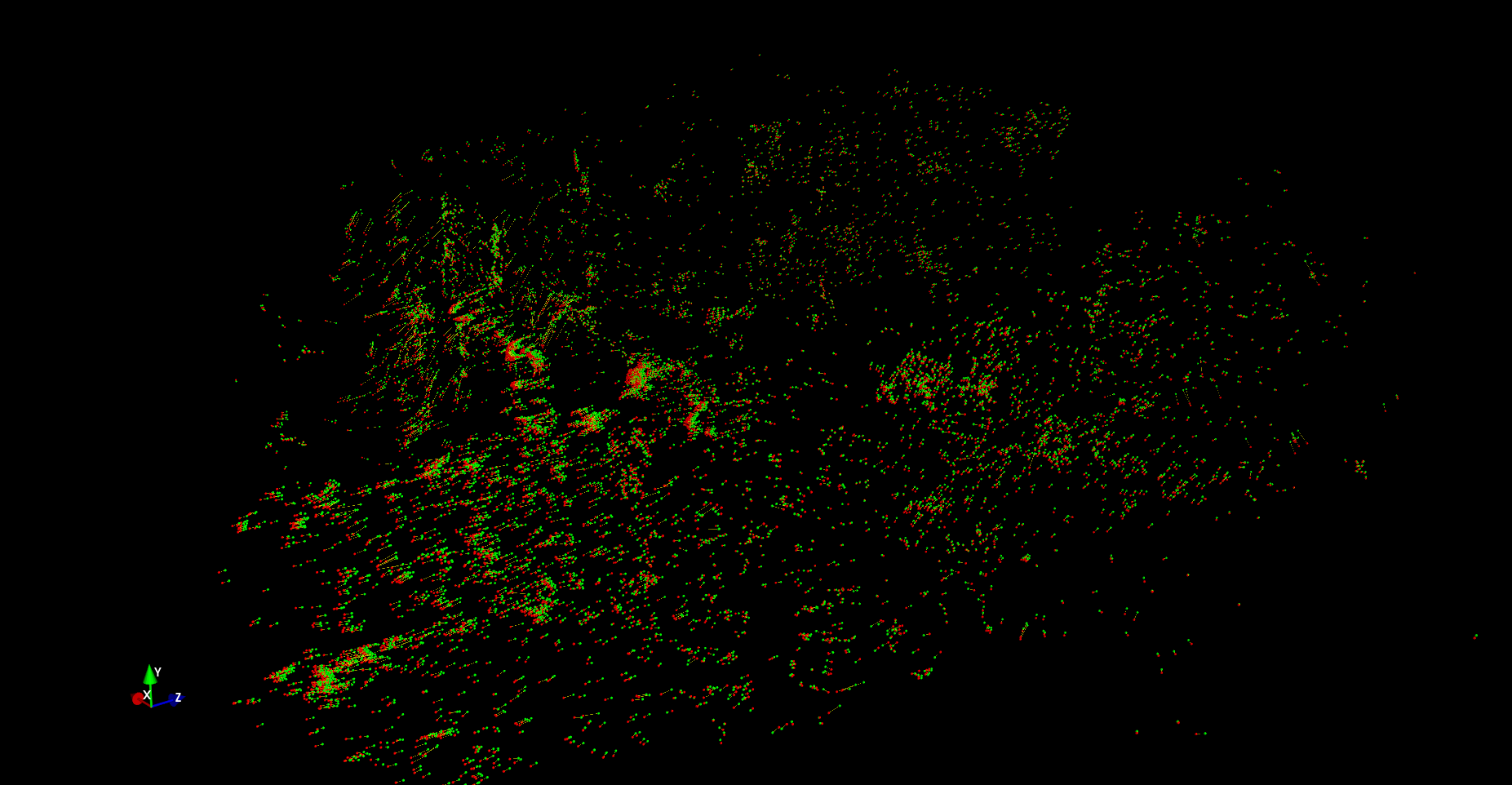}}{}
    \caption{
    The impact of PGD-SF on a sample of point clouds from KITTI.
    The left image shows an example of a point cloud, and the right image illustrates the difference between unattacked and attacked scene flow predictions. The red and green points represent $PC_1 + SF$ before and after PGD-SF, respectively.
    }
    \label{Figure:hplflownet}
\end{figure*}

Table \ref{table:table1} and Table \ref{table:table2} show the impact of FGSM-SF and PGD-SF attacks on different scene flow networks, including
FlowNet3D \cite{b4}, FLOT \cite{b3}, and HPLFlowNet \cite{b2}. These networks leverage positional information from the points within the point clouds for the scene flow estimation, which is why the attacks specifically targeted the dimensions. As observed in the tables, average end-point error (AEPE) was measured for pre-attacks, attacks per dimension, and attacks across all dimensions. Relative degradation in AEPE is also provided. All the perturbations were scaled such that the norm value $\epsilon = 2$. For this specific setting, a total of 10 iterations were chosen for the PGD-SF. Note that $\alpha = \frac{2.5 \cdot \epsilon}{\text{number of iterations}}$, following the conventions outlined in \cite{pgd3d}. This ensures that the perturbation is capable of reaching the boundary of the $\epsilon$ ball. The results demonstrate that the designed attacks have a significant adverse effect on the performance of all of the models. As indicated in the tables, even attacking a spatial dimension has a substantial impact on increasing the AEPEs. Additionally, the results reveal that PGD-SF is stronger than FGSM-SF, albeit at the cost of longer execution times due to the higher number of steps involved in the algorithm. Among the models, HPLFlowNet is shown to be more robust against adversarial attacks. This may be due to the use of CorrBCLs which behaves similar to the spatial pyramids. 
Fig. \ref{Figure:hplflownet} illustrates the effect of the designed perturbations on HPLFlowNet.

\begin{figure*}[h!]
    \centering
    \small
    \stackunder[2pt]{\includegraphics[width=0.33\textwidth]{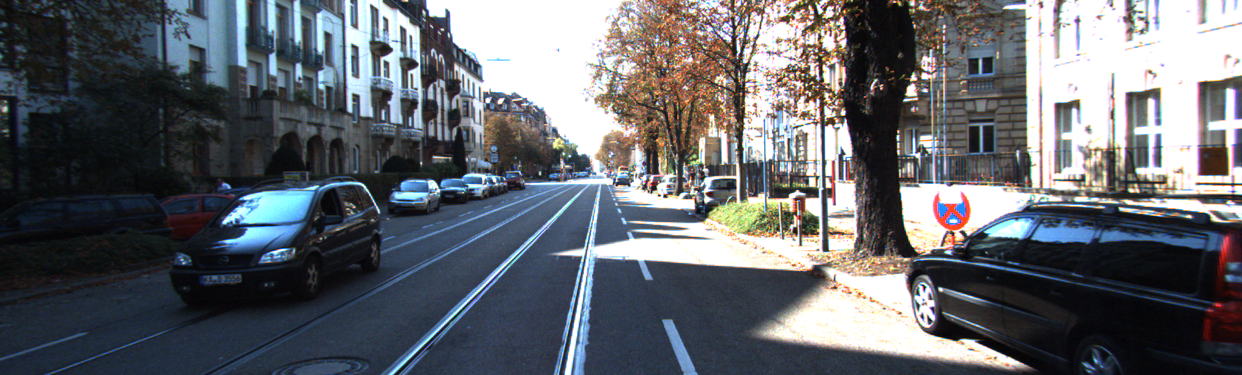}}{Frame 1}  
    \stackunder[2pt]{\includegraphics[width=0.33\textwidth]{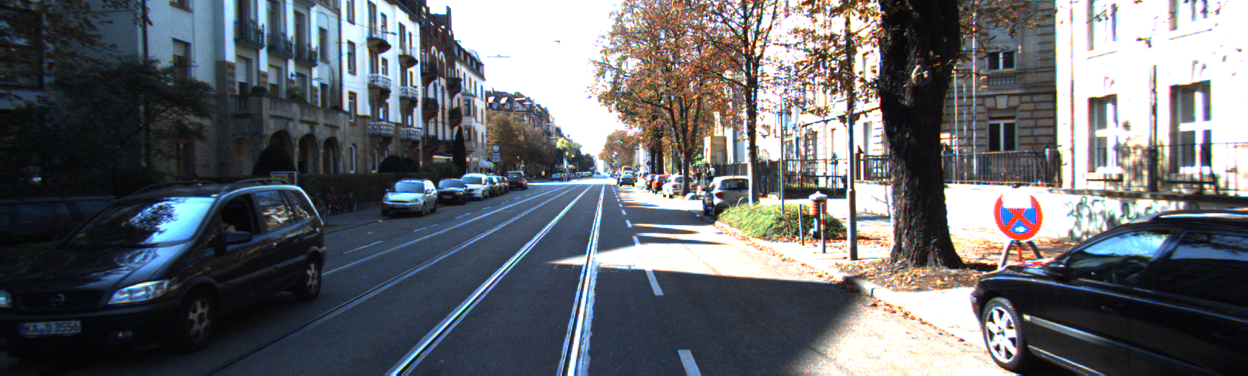}}{Frame 2}  
    \stackunder[2pt]{\includegraphics[width=0.33\textwidth]{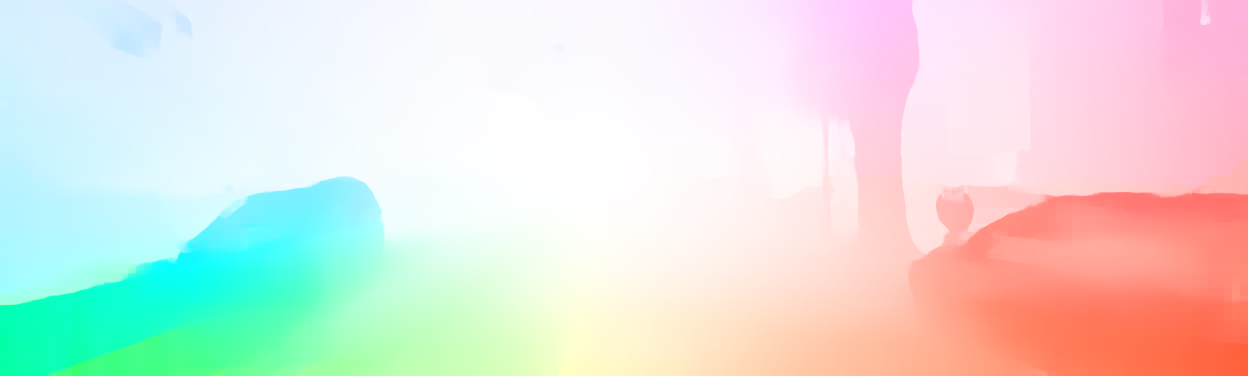}}{Unattacked flow}  

    \vspace{5pt}

    \stackunder[2pt]{\includegraphics[width=0.33\textwidth]{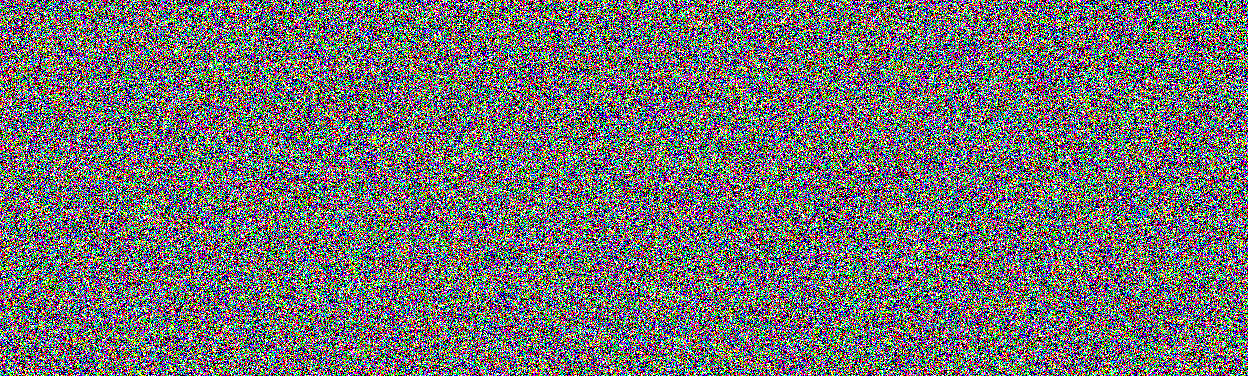}}{Random perturbation}
    \stackunder[2pt]{\includegraphics[width=0.33\textwidth]{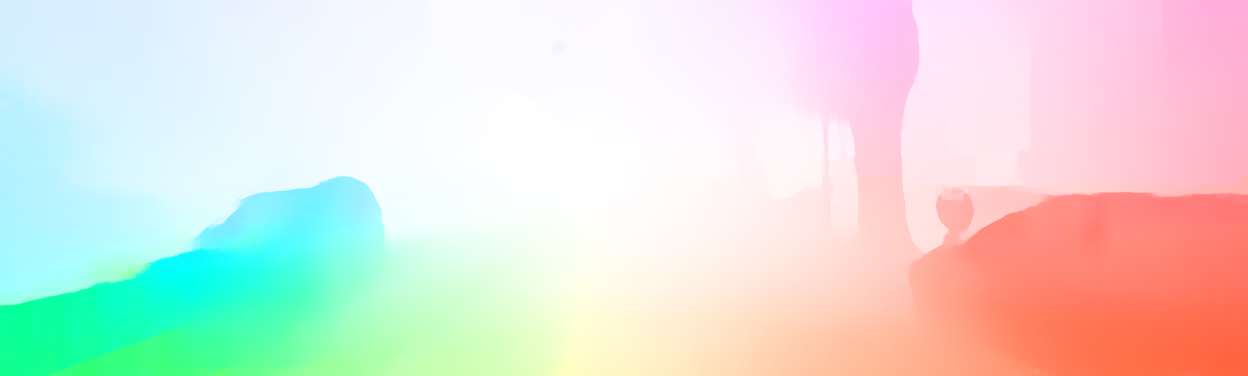}}{Attacked flow} 
    \stackunder[2pt]{\includegraphics[width=0.33\textwidth]{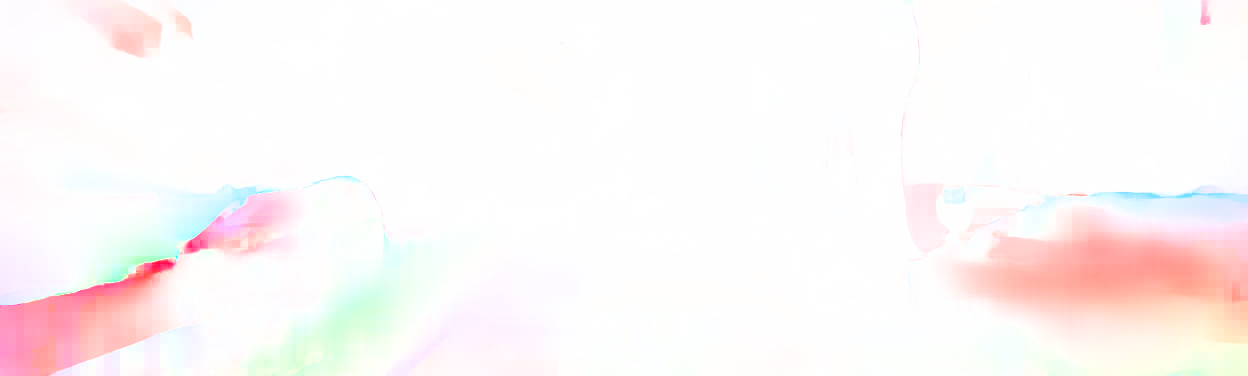}}{Difference} 

    \vspace{5pt}

    \stackunder[2pt]{\includegraphics[width=0.33\textwidth]{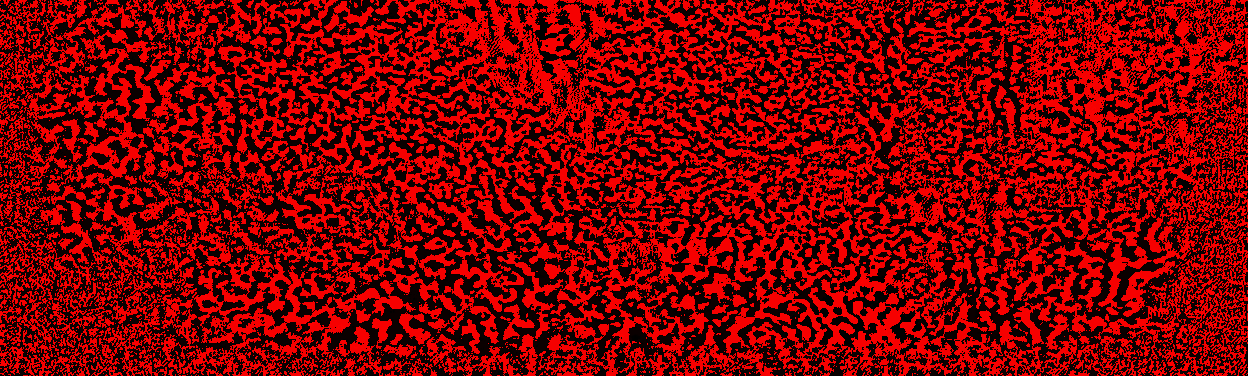}}{PGD-SF perturbation for channel 0}
    \stackunder[2pt]{\includegraphics[width=0.33\textwidth]{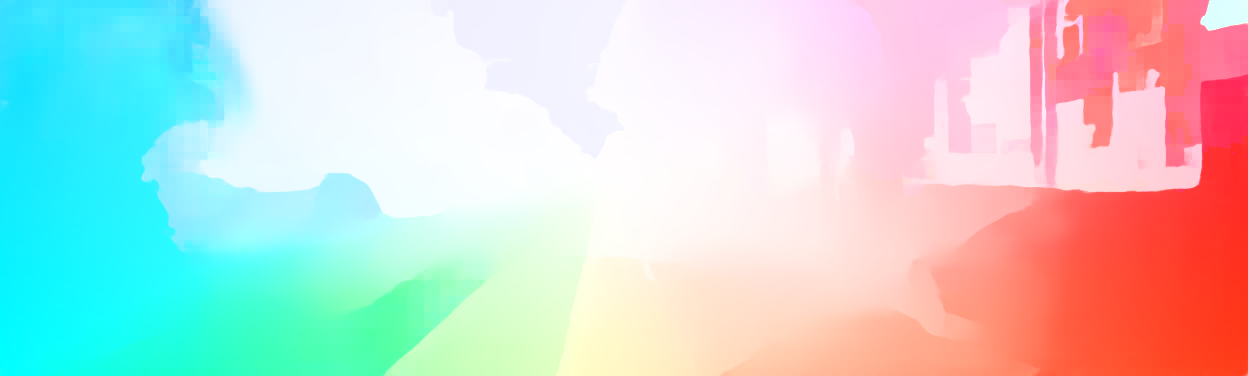}}{Attacked flow}
    \stackunder[2pt]{\includegraphics[width=0.33\textwidth]{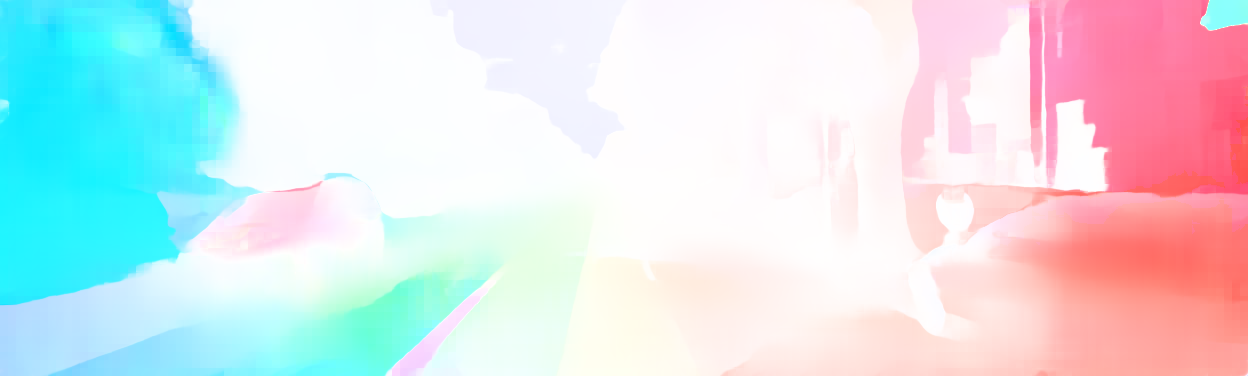}}{Difference} 

    \vspace{5pt}

    \stackunder[2pt]{\includegraphics[width=0.33\textwidth]{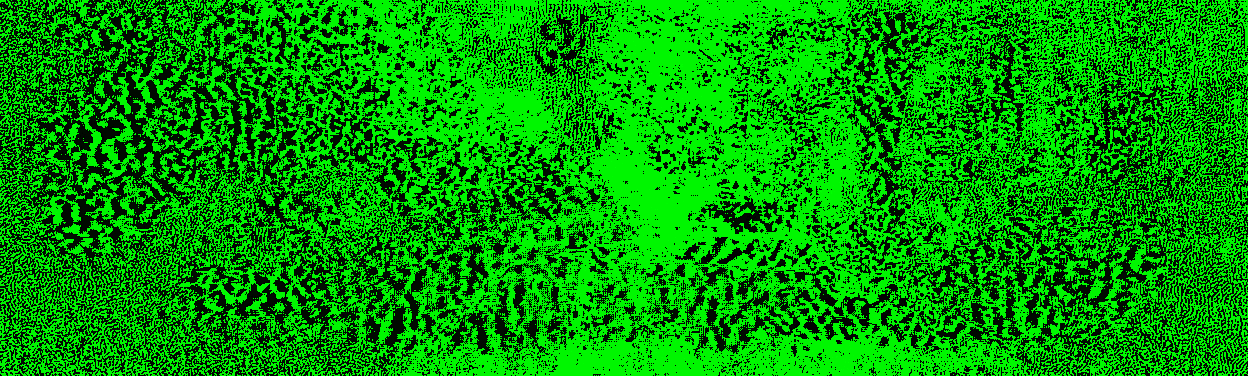}}{PGD-SF perturbation for channel 1}
    \stackunder[2pt]{\includegraphics[width=0.33\textwidth]{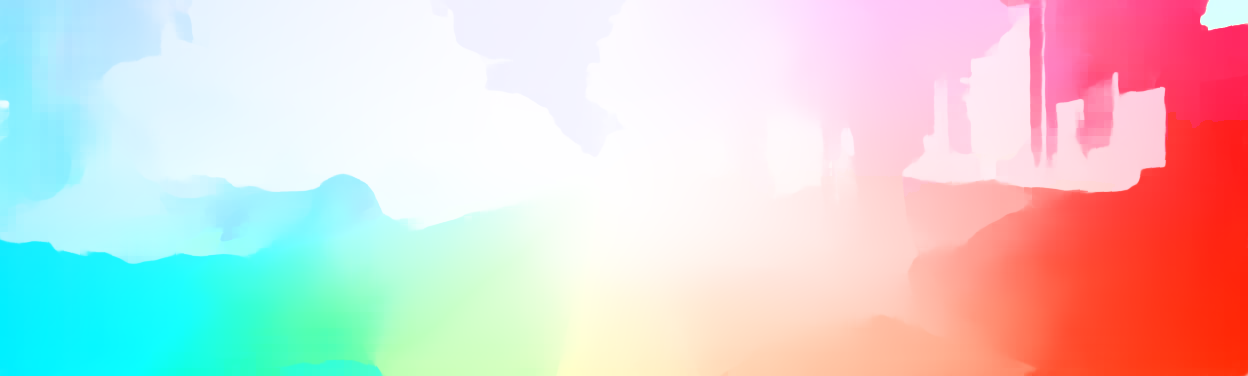}}{Attacked flow}
    \stackunder[2pt]{\includegraphics[width=0.33\textwidth]{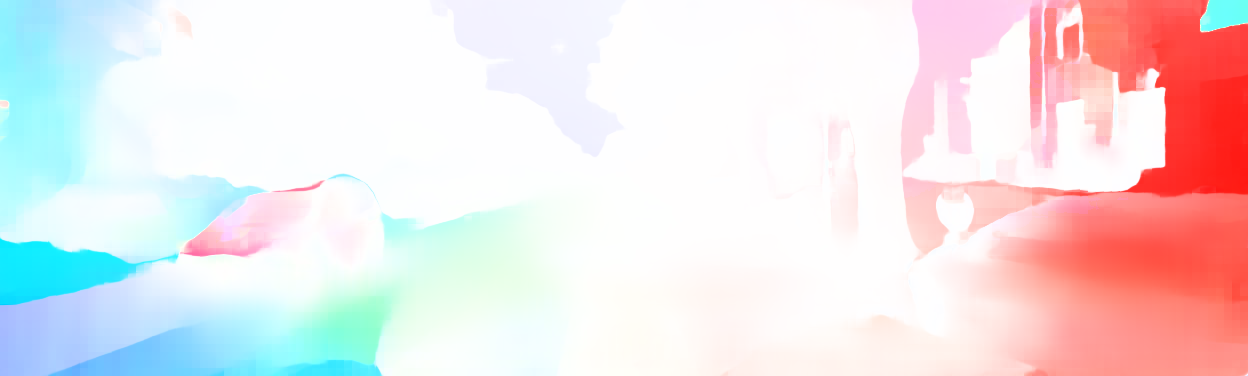}}{Difference} 

    \vspace{5pt}

    \stackunder[2pt]{\includegraphics[width=0.33\textwidth]{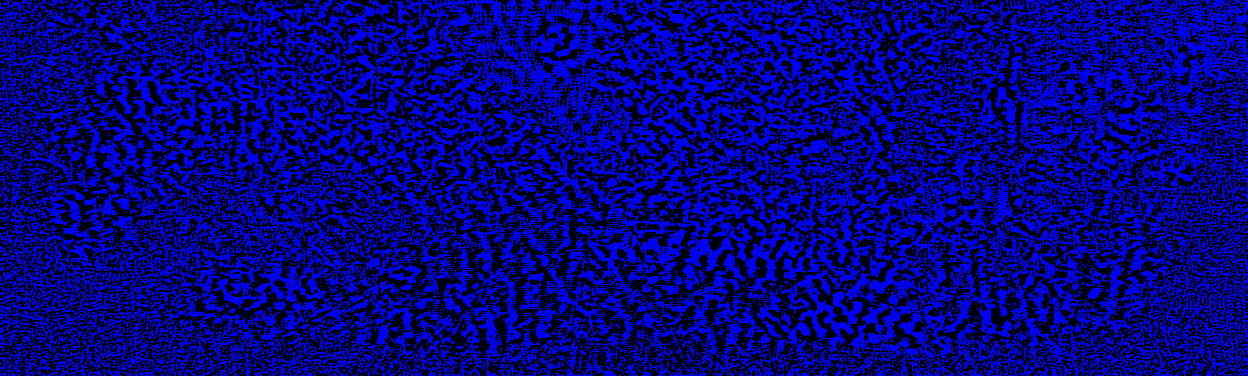}}{PGD-SF perturbation for channel 2}
    \stackunder[2pt]{\includegraphics[width=0.33\textwidth]{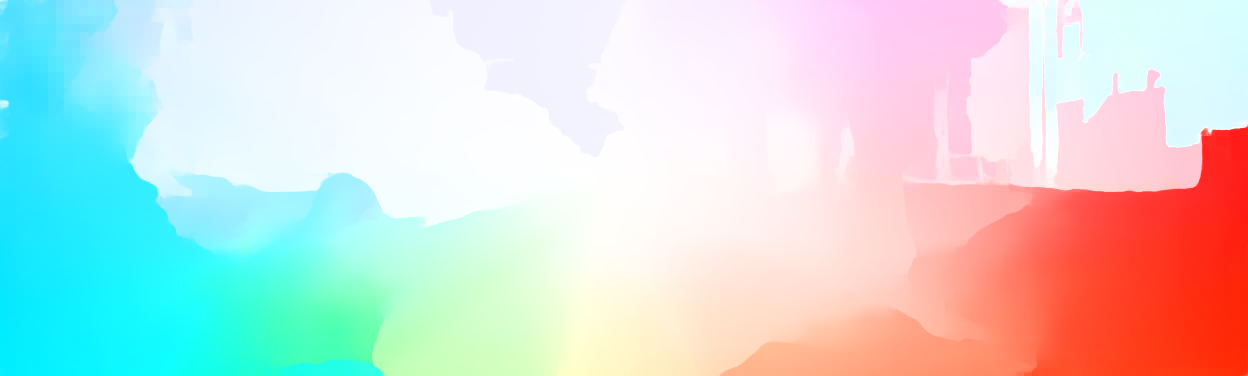}}{Attacked flow}
    \stackunder[2pt]{\includegraphics[width=0.33\textwidth]{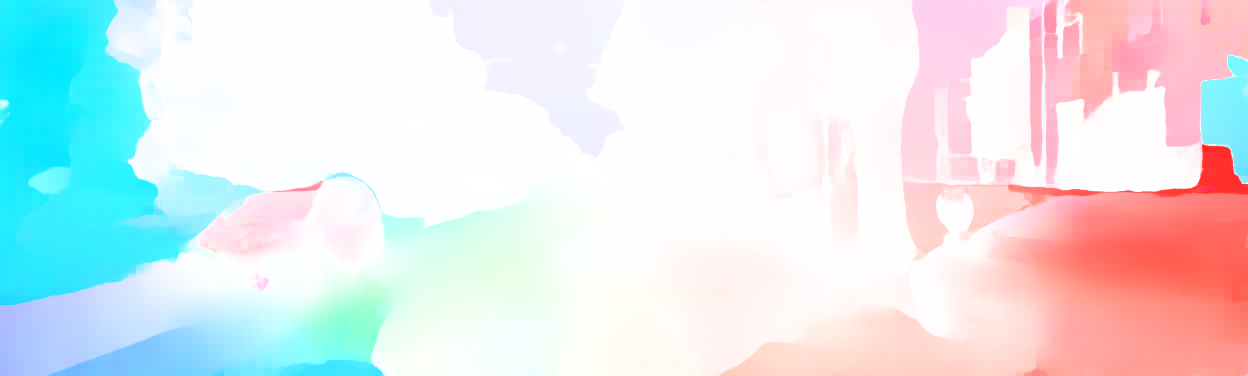}}{Difference} 

    \vspace{5pt}

    \stackunder[2pt]{\includegraphics[width=0.33\textwidth]{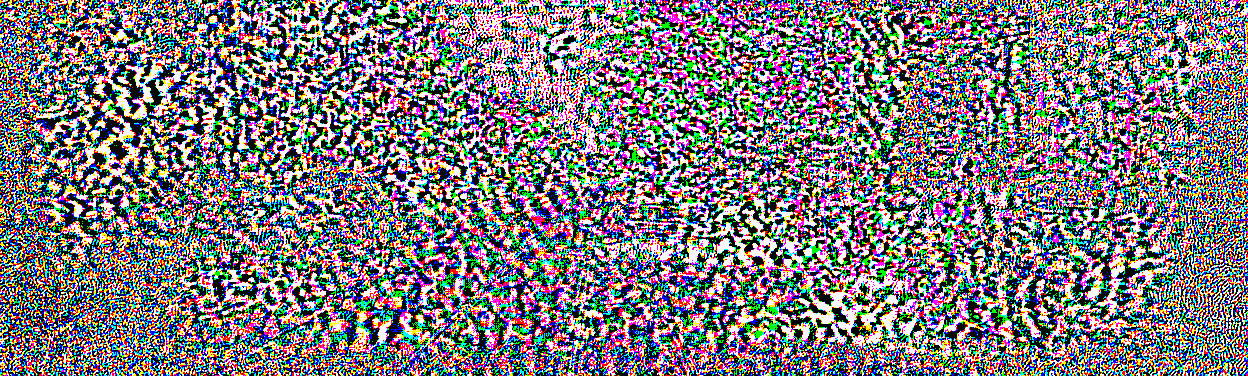}}{PGD-SF perturbation for all channels}
    \stackunder[2pt]{\includegraphics[width=0.33\textwidth]{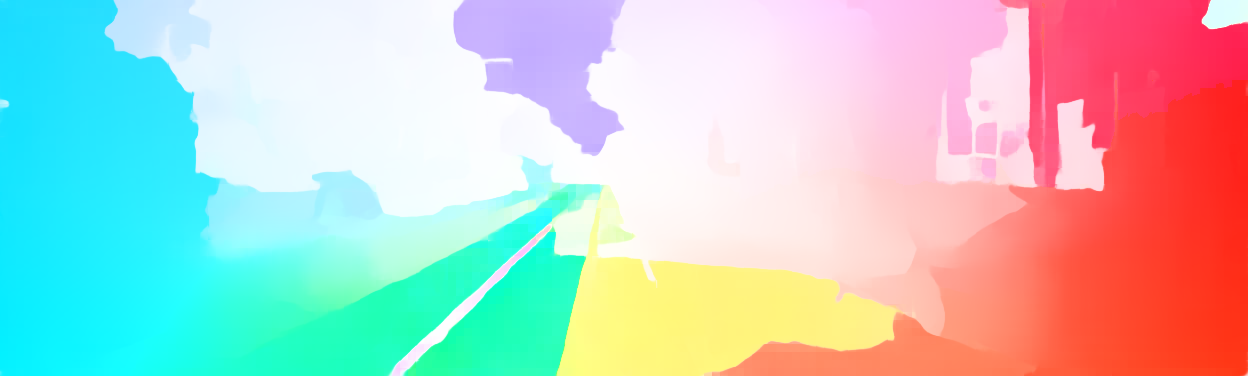}}{Attacked flow} 
    \stackunder[2pt]{\includegraphics[width=0.33\textwidth]{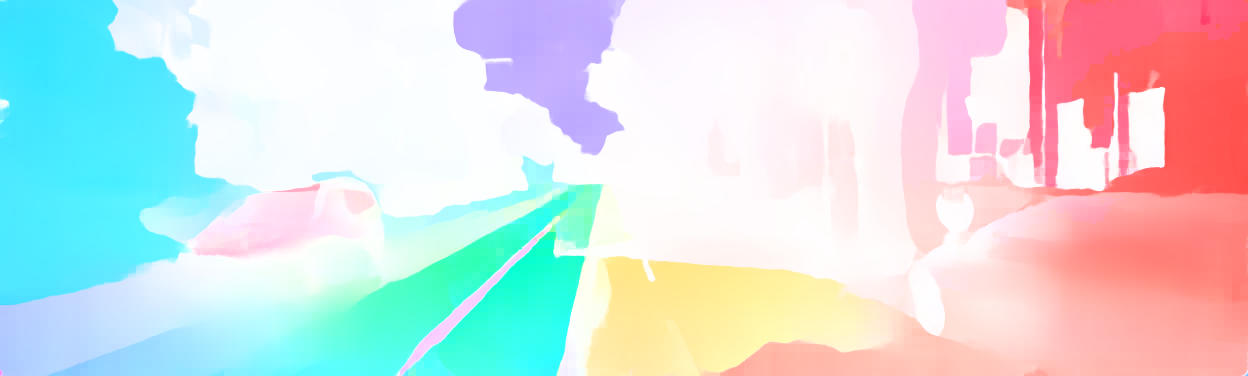}}{Difference} 
    
    \caption{Optical flow of two consecutive frames of KITTI obtained from RAFT before and after attacks.}
    \label{Figure:optical_flow}
\end{figure*}

With many scene flow networks drawing inspiration from their optical flow counterparts, the question arises as to which one exhibits greater robustness. Table \ref{table:table3} compares the robustness of two scene flow networks (RAFT-3D \cite{b5} and FlowNet3D \cite{b4}) with their 2D variants (RAFT \cite{b16} and FlowNet \cite{b17}).
Since all the networks rely on color information for flow estimation, the attacks are applied targeting the color channels. 
In the table, AEPEs are displayed for pre-attacks, attacks per color channel, and attacks across all color channels. Here, the focus was exclusively on evaluating the networks using the KITTI dataset due to its safety-critical applications. For this particular case, a norm value of $\epsilon=\frac{10}{256}$ was selected. It is important to note that although RAFT-3D is categorized as a scene flow network, it differs from the others in that it takes RGB-D data as the input instead of the point clouds. While a direct comparison of AEPEs is not feasible due to the disparity in the number of critical points in an image and the number of points in a point cloud, the relative degradation value facilitates meaningful comparisons by mitigating the influence of point count differences.
It is demonstrated that attacking a single color channel has a significant impact on increasing the AEPEs, particularly showing a more adverse effect for PGD-SF. A good example is Fig. \ref{Figure:optical_flow} which displays the impact of the channel attacks using PGD-SF on optical flow estimation of RAFT.
The results also indicate that the 3D variants of the networks exhibit significantly greater robustness compared to their 2D counterparts. The reason for this lies in the additional information they possess regarding the depth of the points. Upon closer examination of the results, it becomes evident that FlowNet3D demonstrates greater robustness against the channel attacks compared to the dimension attacks. This can be attributed to the significance of the point positions in the scene flow estimation. 
Another intriguing observation is that when incorporating perturbations, the object boundaries become more pronounced within the resulting optical flow. This enhancement can potentially aid in achieving improved segmentation. Additionally, Table \ref{table:table4} demonstrates the effects of random attacks on the discussed networks using KITTI. The value of $\epsilon$ for these attacks is selected following the same settings as discussed earlier. The table clearly indicates that random attacks have no noticeable impact on the average end-point error, in contrast to the proposed attacks. Overall, these empirical results confirm the hypothesis that both FGSM-SF and PGD-SF can significantly change the scene flow.

\begin{table*}[t]
\begin{center}
\caption{Comparison between 2D and 3D networks.}
\label{table:table3}
\resizebox{\textwidth}{!}{%
\begin{tabular}{l|l|c|cc|cc|cc|cc}
\toprule
\multirow{2}{*}{Attack type} & \multirow{2}{*}{Model} & Unattacked & \multicolumn{2}{|c|}{Attack on channel 1} & \multicolumn{2}{|c|}{Attack on channel 2} & \multicolumn{2}{|c|}{Attack on channel 3} & \multicolumn{2}{|c}{Attack on all channels} \\
{} & {} & AEPE & AEPE & Rel & AEPE & Rel & AEPE & Rel & AEPE & Rel \\
\midrule
\multirow{4}{*}{FGSM-SF} & RAFT & 5.030 & 8.120 & +6.579 & 5.740 & +0.141 & 6.330 & +0.258 & 8.830 & +0.755 \\
{} & RAFT-3D & 11.158 & 10.962 & -0.018 & 11.193 & +0.003 & 10.851 & -0.028 & 10.626 & -0.048 \\
\cmidrule{2-11}
{} & FlowNet & 7.789 & 38.800 & +3.981 & 27.222 & +2.495 & 41.455 & +4.322 & 70.193 & +8.012 \\
{} & FlowNet3D & 0.292 & 0.896 & +2.068 & 1.179 & +3.038 & 1.154 & +2.952 & 1.115 & +2.818 \\
\midrule
\multirow{4}{*}{PGD-SF} & RAFT & 5.030 & 22.160 & +3.406 & 10.495 & +1.086 & 13.502 & +1.684 & 37.098 & +6.375 \\
{} & RAFT-3D & 11.158 & 11.758 & +0.054 & 11.795 & +0.057 & 11.826 & +0.060 & 11.986 & +0.074 \\
\cmidrule{2-11}
{} & FlowNet & 7.789 & 97.247 & +11.485 & 76.563 & +8.830 & 93.250 & +10.972 & 334.929 & +42.000 \\
{} & FlowNet3D & 0.292 & 0.928 & +2.178 & 1.324 & +3.534 & 1.196 & +3.096 & 1.358 & +3.651 \\
\bottomrule
\end{tabular}
}
\end{center}
\end{table*}

\begin{table}
\begin{center}
\caption{Impact of random attacks on
average end-point error.}
\label{table:table4}
\resizebox{0.55\columnwidth}{!}{%
\begin{tabular}{c|c|cc|c}
\toprule
\multirow{2}{*}{Model} & Unattacked & \multicolumn{2}{|c|}{Random attack} & \multirow{2}{*}{Target} \\
{} & AEPE & AEPE & Rel & {}\\
\midrule
HPLFlowNet & 0.117 & 0.119 & +0.017 & Dimension \\
\midrule
FLOT & 0.110 & 2.377 & +20.609 & Dimension \\
\midrule
FlowNet & 7.789 & 11.332 & +0.455 & Color channel\\
\midrule
\multirow{2}{*}{FlowNet3D} &  \multirow{2}{*}{0.292} & 2.378 & +7.144 & Dimension \\
{} & {} & 0.965 & +2.305 & Color channel \\
\midrule
RAFT & 5.030 & 5.305 & +0.055 & Color channel\\
\midrule
RAFT-3D & 11.158 & 11.174 & +0.001 & Color channel \\
\bottomrule
\end{tabular}
}
\end{center}
\end{table}

\section{Conclusion}
\label{sec:conclusion}

A novel adversarial attack was introduced to deceive scene flow networks while preserving human perceptual consistency. The generated adversarial examples effectively increased the average end-point error of the targeted models. The susceptibility of the scene flow networks to the adversarial attacks was observed to be higher compared to the optical flow networks. This highlights the relative under-exploration of adversarial robustness in scene flow estimation and the need for more comprehensive study in this area. Beyond evaluation, the proposed attack can also be used within adversarial training frameworks to improve model robustness. Given the importance of scene flow estimation in applications such as liveness detection and authentication, future work could investigate integrating adversarial robustness methods into these systems. Another promising direction is the development of black-box and universal adversarial attacks for scene flow estimation.

\vfill\pagebreak
\clearpage


\bibliographystyle{IEEEbib}
\bibliography{strings,refs}

@InProceedings{b2,
  title={HPLFlowNet: Hierarchical Permutohedral Lattice FlowNet for
Scene Flow Estimation on Large-scale Point Clouds},
  author={Gu, Xiuye and Wang, Yijie and Wu, Chongruo and Lee, Yong Jae and Wang, Panqu},
  booktitle={Computer Vision and Pattern Recognition (CVPR), 2019 IEEE International Conference on},
  year={2019}
}

@article{b7,
  title={FlowStep3D: Model Unrolling for Self-Supervised Scene Flow Estimation},
  author={Yair Kittenplon and Yonina C. Eldar and Dan Raviv},
  journal={ArXiv},
  year={2020},
  volume={abs/2011.10147}
}

@InProceedings{b3,
  title={{FLOT}: {S}cene {F}low on {P}oint {C}louds {G}uided by {O}ptimal {T}ransport},
  author={Puy, Gilles and Boulch, Alexandre and Marlet, Renaud},
  booktitle={European Conference on Computer Vision},
  year={2020}
}

@article{b4,
      title={FlowNet3D: Learning Scene Flow in 3D Point Clouds},
      author={Liu, Xingyu and Qi, Charles R and Guibas, Leonidas J},
      journal={CVPR},
      year={2019}
}

@InProceedings{b5,
  title={RAFT-3D: Scene Flow using Rigid-Motion Embeddings},
  author={Teed, Zachary and Deng, Jia},
  booktitle={Proceedings of the IEEE/CVF Conference on Computer Vision and Pattern Recognition (CVPR)},
  year={2021},
}

@InProceedings{b9,
  title={PointPWC-Net: Cost Volume on Point Clouds for (Self-) Supervised Scene Flow Estimation},
  author={Wu, Wenxuan and Wang, Zhi Yuan and Li, Zhuwen and Liu, Wei and Fuxin, Li},
  booktitle={European Conference on Computer Vision},
  pages={88--107},
  year={2020},
  organization={Springer}
}

@InProceedings{b6,
  author={Liu, Daniel and Yu, Ronald and Su, Hao},
  booktitle={2019 IEEE International Conference on Image Processing (ICIP)}, 
  title={Extending Adversarial Attacks and Defenses to Deep 3D Point Cloud Classifiers}, 
  year={2019},
  volume={},
  number={},
  pages={2279-2283},
  doi={10.1109/ICIP.2019.8803770}
}

@InProceedings{b8,
title	= {Intriguing properties of neural networks},
author	= {Christian Szegedy and Wojciech Zaremba and Ilya Sutskever and Joan Bruna and Dumitru Erhan and Ian Goodfellow and Rob Fergus},year	= {2014},
URL	= {http://arxiv.org/abs/1312.6199},
booktitle	= {International Conference on Learning Representations}
}

@InProceedings{
b10,
title={Towards Deep Learning Models Resistant to Adversarial Attacks},
author={Aleksander Madry and Aleksandar Makelov and Ludwig Schmidt and Dimitris Tsipras and Adrian Vladu},
booktitle={International Conference on Learning Representations},
year={2018},
url={https://openreview.net/forum?id=rJzIBfZAb},
}

@article{b1,
  title={Attacking Optical Flow},
  author={Anurag Ranjan and Joel Janai and Andreas Geiger and Michael J. Black},
  journal={2019 IEEE/CVF International Conference on Computer Vision (ICCV)},
  year={2019},
  pages={2404-2413},
  url={https://api.semanticscholar.org/CorpusID:204823874}
}

@article{b11,
  title={Universal Adversarial Training},
  author={Ali Shafahi and Mahyar Najibi and Zheng Xu and John P. Dickerson and Larry S. Davis and Tom Goldstein},
  journal={ArXiv},
  year={2018},
  volume={abs/1811.11304},
  url={https://api.semanticscholar.org/CorpusID:53806548}
}

@InProceedings{b12,
  title={Universal adversarial perturbations},
  author={Seyed-Mohsen Moosavi-Dezfooli* and Alhussein Fawzi* and Omar Fawzi and Pascal Frossard},
  journal={CVPR},
  year={2017},
}

@article{b13,
  title={Interpretation of Neural Networks is Susceptible to Universal Adversarial Perturbations},
  author={Haniyeh Ehsani Oskouie and Farzan Farnia},
  journal={ICASSP 2023 - 2023 IEEE International Conference on Acoustics, Speech and Signal Processing (ICASSP)},
  year={2022},
  pages={1-5},
  url={https://api.semanticscholar.org/CorpusID:254274969}
}

@INPROCEEDINGS{b14,
  author={Geiger, Andreas and Lenz, Philip and Urtasun, Raquel},
  booktitle={2012 IEEE Conference on Computer Vision and Pattern Recognition}, 
  title={Are We Ready for Autonomous Driving? The KITTI Vision Benchmark Suite}, 
  year={2012},
  volume={},
  number={},
  pages={3354-3361},
  doi={10.1109/CVPR.2012.6248074}}

@INPROCEEDINGS{b15,
  author={Mayer, Nikolaus and Ilg, Eddy and Häusser, Philip and Fischer, Philipp and Cremers, Daniel and Dosovitskiy, Alexey and Brox, Thomas},
  booktitle={2016 IEEE Conference on Computer Vision and Pattern Recognition (CVPR)}, 
  title={A Large Dataset to Train Convolutional Networks for Disparity, Optical Flow, and Scene Flow Estimation}, 
  year={2016},
  volume={},
  number={},
  pages={4040-4048},
  doi={10.1109/CVPR.2016.438}
}

@inproceedings{b16,
  title={RAFT: Recurrent All-Pairs Field Transforms for Optical Flow},
  author={Zachary Teed and Jia Deng},
  booktitle={European Conference on Computer Vision},
  year={2020},
  url={https://api.semanticscholar.org/CorpusID:214667893}
}

@article{b17,
  title={FlowNet: Learning Optical Flow with Convolutional Networks},
  author={Alexey Dosovitskiy and Philipp Fischer and Eddy Ilg and Philip H{\"a}usser and Caner Hazirbas and Vladimir Golkov and Patrick van der Smagt and Daniel Cremers and Thomas Brox},
  journal={2015 IEEE International Conference on Computer Vision (ICCV)},
  year={2015},
  pages={2758-2766},
  url={https://api.semanticscholar.org/CorpusID:12552176}
}

@ARTICLE{attack,
  author={Naderi, Hanieh and Bajić, Ivan V.},
  journal={IEEE Access}, 
  title={Adversarial Attacks and Defenses on 3D Point Cloud Classification: A Survey}, 
  year={2023},
  volume={11},
  number={},
  pages={144274-144295},
  doi={10.1109/ACCESS.2023.3345000}}

@inproceedings{
pgd3d,
title={Towards Deep Learning Models Resistant to Adversarial Attacks},
author={Aleksander Madry and Aleksandar Makelov and Ludwig Schmidt and Dimitris Tsipras and Adrian Vladu},
booktitle={International Conference on Learning Representations},
year={2018},
url={https://openreview.net/forum?id=rJzIBfZAb},
}

\end{document}